\documentclass{llncs}

\usepackage{graphicx}
\usepackage{amsmath}
\usepackage{latexsym}
\usepackage{amssymb}
\usepackage{amsfonts}
\usepackage{url}
\usepackage{comment}
\usepackage[boxed,noend]{algorithm2e}
\usepackage{xspace}

\newcommand{\A}{\mathcal{A}} 
 
 \newcommand{\F}{\mathcal{F}}

\newcommand{\K}{\mathcal{K}} \renewcommand{\L}{\mathcal{L}}

 \newcommand{\R}{\mathcal{R}}
\renewcommand{\S}{\mathcal{S}} \newcommand{\T}{\mathcal{T}}
\newcommand{\U}{\mathcal{U}}

\newcommand{\lora}{\longrightarrow}

\newcommand{\tup}[1]{\langle #1\rangle}            

\newcommand{\NOT}{\neg}

\newcommand{\SOMET}[1]{\exists #1}

\newcommand{\INV}[1]{#1^{-}}

\newcommand{\ISA}{\sqsubseteq}

\newcommand{\dllite}{\textit{DL-Lite}\xspace}

\newcommand{\dllitefr}{\textit{DL-Lite}_{F\R}\xspace}

\newcommand{\dlliteaid}{\textit{DL-Lite}_{A,id}\xspace}

\newcommand{\ALCQIO}{\mathcal{ALCQIO}}

\newcommand{\funct}[1]{(\mathsf{funct}\; #1)}

\newcommand{\DOMAIN}[1]{\delta(#1)}
\newcommand{\RANGE}[1]{\rho(#1)}

\newcommand{\clos}[2]{\textsf{cl}_{#1}(#2)}

\newcommand{\Mod}[1]{\mathit{Mod}{(#1)}}

\newcommand{\FOT}{FT}

\def\qed{\hfill{\qedboxempty}      
  \ifdim\lastskip<\medskipamount \removelastskip\penalty55\medskip\fi}

\def\qedboxempty{\vbox{\hrule\hbox{\vrule\kern3pt
                 \vbox{\kern3pt\kern3pt}\kern3pt\vrule}\hrule}}

\def\qedfull{\hfill{\qedboxfull}   
  \ifdim\lastskip<\medskipamount \removelastskip\penalty55\medskip\fi}

\def\qedboxfull{\vrule height 4pt width 4pt depth 0pt}

\renewenvironment{proof}{\textsl{Proof.\ }}{\qedfull}
\newenvironment{proofsk}{\textsl{Proof (sketch).\ }}{\qedfull}

\title{On the evolution of the instance level of $\dllite$
  knowledge bases}
\author{Maurizio Lenzerini, Domenico Fabio Savo}

\institute{Dipartimento di Informatica e Sistemistica\\
   \textsc{Sapienza} Universit\`a di Roma\\
   \texttt{\textit{lastname}@dis.uniroma1.it}
}

\begin{document}

\maketitle

\begin{abstract}
  Recent papers address the issue of updating the instance level of
  knowledge bases expressed in Description Logic following a
  model-based approach. One of the outcomes of these papers is that
  the result of updating a knowledge base $\K$ is generally not
  expressible in the Description Logic used to express $\K$. In this
  paper we introduce a formula-based approach to this problem, by
  revisiting some research work on formula-based updates developed in
  the '80s, in particular the WIDTIO (When In Doubt, Throw It Out)
  approach. We show that our operator enjoys desirable properties,
  including that both insertions and deletions according to such
  operator can be expressed in the DL used for the original KB.  Also,
  we present polynomial time algorithms for the evolution of the
  instance level knowledge bases expressed in $\dlliteaid$, which the
  most expressive Description Logics of the $\dllite$ family.
\end{abstract}

\section{Introduction}\label{sec:introduction}

Description Logics (DLs) \cite{BCMNP03} are logics for expressing
knowledge bases (KBs) constituted by two components, namely, the TBox,
asserting general properties of concepts and roles (binary relations),
and the ABox, which is a set of assertions about individuals that are
instances of concepts and roles. It is widely accepted that such
logics are well-suited for expressing ontologies, with the TBox
capturing the intensional knowledge about the domain of interest, and
the ABox expressing the knowledge about the instance level of the
predicates defined in the TBox. Following this idea, several Knowledge
Representation Systems, called DL systems, have been recently built,
providing methods and tools for managing ontologies expressed in
DLs~\footnote{http://www.cs.man.ac.uk/~sattler/reasoners.html}. 
Notice that numerous DLs have been studied in the last decades, with
the goal of analyzing the impact of the expressive power of the DL
language to the complexity of reasoning. Consequently, each DL system
is tailored towards managing KB expressed in a specific DL.

By referring to the so-called \emph{functional view of knowledge
  representation}~\cite{Leve84}, DL systems should be able to
perform two kinds of operations, called {\sf ASK} and {\sf
TELL}. {\sf
  ASK} operations, such as subsumption checking, or query answering,
are used to extract information from the KB, whereas {\sf
TELL} operations aim at changing the KB according to new
knowledge acquired over the domain. In other words, {\sf TELL}
operations should be able to cope with the \emph{evolution of the
KB}.

There are two types of evolution operators, corresponding to
inserting, and deleting chunks of knowledge, respectively. In the case
of insertion, the aim is to incorporate new knowledge into the KB, and
the corresponding operator should be defined in such a way to compute
a consistent KB that supports the new knowledge. In the case of
deletion, the aim is to come up with a consistent KB where the
retracted knowledge is not valid. In both cases, the crucial aspect to
take into account is that evolving a consistent knowledge base should
not introduce inconsistencies. We point out that a different approach
would be to allow inconsistencies in the KB, and then resorting to
sophisticated quesy answering mechanisms, tolerant to such
inconsistencies (see, for example, \cite{ArBC99,LEFF*05}, but this is
outside the scope of the approach presented here.

Notice that, while {\sf ASK} operations have been investigated in
detail by the DL community, existing DL reasoners do not provide
explicit services for KB evolution. Nevertheless, many recent papers
demonstrate that the interest towards a well-defined approach to KB
evolution is growing
significantly~\cite{FMKPA08,LLMW06,DLPR09,WangWT10,CKNZ10b}.

Following the tradition of the work on knowledge revision and
update~\cite{KaMe91}, all the above papers advocate some minimality
criterion in the changes of the KB that must be undertaken to realize
the evolution operations. In other words, the need is commonly
perceived of keeping the distance between the original KB and the KB
resulting from the application of an evolution operator minimal. There
are two main approaches to define such a distance, called
\emph{model-based} and \emph{formula-based}, respectively. In the
model-based approaches, the result of an evolution operation applied
to the KB $\K$ is defined in terms of a set of models, with the idea
that such a set should be as close as possible to the models of $\K$.
One basic problem with this approach is to characterize the language
needed to express the KB that exactly captures the resulting set of
models. Conversely, in the formula-based approaches, the result is
explicitly defined in terms of a formula, by resorting to some
minimality criterion with respect to the formula expressing
$\K$. Here, the basic problem is that the formula constituting the
result of an evolution operation is not unique in general.

In this paper, we study the problem of DL KB evolution, by focusing
our attention to scenarios characterized by the following elements:
\begin{enumerate}
\item We consider the case where the evolution affects only the
  instance level of the KB, i.e., the ABox. In other words, we enforce
  the condition that the KB resulting from the application of the
  evolution operators has the same TBox as the original KB (similarly
  to \cite{LLMW06,DLPR09}). 
\item We aim at a situation where the KB resulting from the evolution
  can be expressed in the same DL as the original KB. This is coherent
  with our goal of providing the foundations for equipping DL systems
  with evolution operators: indeed, if a DL system $S$ is able to
  manage KBs expressed in a DL $\L$, the result of evolving such KBs
  should be expressible in $\L$.
\item The KBs resulting from the application of an evolution operator
  on two logically equivalent KBs should be mutually equivalent. In
  other words, we want the result to be independent of the syntactic
  form of the original KB.
\end{enumerate}
Assumption (1), although limiting the generality of our approach,
captures several interesting scenarios, including \emph{ontology-based
  data management}, where the DL KB is used as a logic-based interface
to existing information systems (databases, web sources, etc.).

As for item (2), we note that virtually all model-based
approaches suffer from the expressibility problem. This has
been reported in many recent papers, including
\cite{LLMW06,DLPR09,CKNZ10b}, 
for DLs whose expressive power range from \dllite\ to
$\ALCQIO$. For this reason, we adopt a formula-based approach,
inspired in particular by the work developed in \cite{FaUV83}
for updating logical theories. As in \cite{FaUV83}, we
consider both insertions and deletions. However, we differ
from \cite{FaUV83} for an important aspect. We already noted
that the formula constituting the result of an evolution
operation is not unique in general. While \cite{FaUV83}
essentially proposes to keep the whole set of such formulas,
we take a radical approach, and consider their intersection as
the result of the evolution. In other words, we follow the
\emph{When In Doubt Throw It Out} (WIDTIO)
\cite{GiSm87,Wins90} principle.

Finally, to deal with item (3), we sanction that the notion of
distance between KBs refers to the closure of the ABox of a KB, rather
than to the ABox itself. The closure of an ABox $\A$ with respect to
an TBox $\T$ is defined as the set of all ABox assertions that
logically follows from $\T$ and $\A$. By basing the definition of
distance on the closure of ABoxes, we achieve the goal of making the
result of our operators independent of the form of the original KB.

After a brief introduction to DLs (Section 2), we provide the
definition of our evolution operators in Section 3, together
with a comparison with related approaches. The remaining
sections are devoted to illustrating algorithms for deletion
(Section 4), and insertion (Section 5) for KBs expressed in
the DL $\dlliteaid$, which is the most expressive logic in the
$\dllite$ family~\cite{CDLLR07} 
The $\dllite$ family\footnote{Not to be confused with the set
of DLs studied in~\cite{ACKZ09}, which form the
$\dllite_{bool}$ family.} has been specifically designed to
keep all reasoning tasks polynomially tractable, and we show
that this property still holds for the evolution operators
proposed in this paper. Indeed, we show that computing the
result of both insertions and deletions to KBs expressed
$\dlliteaid$ is tractable.

\section{Preliminaries}\label{sec:language}

Let $\S$ be a signature of symbols for individual (object and value)
constants, and atomic elements, i.e., concepts, value-domains,
attributes, and roles. If $\L$ is a DL, then an $\L$-KB $\K$ over $\S$
is a pair $\tup{\T,\A}$~\cite{BCMNP03} where $\T$, called \emph{TBox},
is a finite set of intensional assertions over $\S$ expressed in $\L$,
and $\A$, called \emph{ABox}, is a finite set of instance assertions,
i.e, assertions on individuals, over $\S$.  Different DLs allow for
different kinds of TBox and/or ABox assertions. In this paper we
assume that ABox assertions are always \emph{atomic}, i.e., they
correspond to ground atoms, and therefore we omit to refer to $\L$
when we talk about ABox assertions.

The semantics of a DL KB is given in terms of first-order
interpretations~\cite{BCMNP03}. An interpretation is a \emph{model} of
a DL knowledge base $\K=\tup{\T,\A}$ if it satisfies all assertions in
$\T \cup \A$ (the notion of satisfaction depends on the constructs
allowed by the specific DL in which $\K$ is expressed). We denote the
set of models of $\K$ with $\Mod\K$.

Let $\T$ be a TBox in $\L$, and let $\A$ be an ABox. We say that $\A$
is \emph{$\T$-consistent} if $\langle \T, \A \rangle$ is satisfiable, i.e. if
$Mod(\tup{\T,\A})\neq \emptyset$, $\T$-inconsistent otherwise.
The \emph{$\T$-closure} of $\A$ with respect to $\T$, denoted
$\clos{\T}{\A}$, is the set of all atomic ABox assertion that are
formed with individuals in $\A$, and are logically implied by $\langle
\T, \A \rangle$. Obviously, $\langle \T,\A \rangle$ is logically
equivalent to $\langle \T,\clos{\T}{\A} \rangle$. $\A$ is said to be
\emph{$\T$-closed} if $\clos{\T}{\A} = \A$. Finally, for an ABox
assertion $\gamma_1$, we denote by $\textsf{Subsumee}_{\langle \T, \A
  \rangle}(\gamma_1)$ the set of atoms $\gamma_2 \in \clos{\T}{\A}$
such that $\langle \T, \A \rangle \models \gamma_2 \supset \gamma_1$.

\noindent \textbf{The description logic $\dlliteaid$.} \; The
$\dllite$ family~\cite{CDLLR07} is a family of low complexity DLs
particularly suited for dealing with KBs with very large ABoxes, and
forms the basis of OWL~2~QL, one of the profile of OWL~2, the official
ontology specification language of the World-Wide-Web Consortium
(W3C)\footnote{
  \url{http://www.w3.org/TR/2008/WD-owl2-profiles-20081008/}}.

We now present the DL $\dlliteaid$, which is the most exprressive
logic in the family. Expressions in $\dlliteaid$ are formed according
to the following syntax:
\[
\centering
  \begin{array}{@{}r@{~}c@{~}l@{\qquad}l@{~}c@{~}l}
    B &\lora&  A  ~\mid~ \SOMET{Q} ~\mid~ \DOMAIN{U} & E &\lora& \RANGE{U}\\
    C &\lora& B ~\mid~ \neg B & T &\lora& \top_D ~\mid~ T_1 ~\mid~ \cdots ~\mid~ T_n\\
    Q &\lora& P ~\mid~\INV{P} &  V &\lora& U ~\mid~ \neg U\\
    R &\lora& Q ~\mid~ \neg Q
    \end{array}
\]
where $A$, $P$, and $U$ are symbols in $\S$ denoting respectively an
atomic concept name, an atomic role name and an attribute name,
$T_1,\ldots,T_n$ are all the value-domains allowed in the logic (those
corresponding to the data types adopted by Resource Description
Framework (RDF)\footnote{\url{http://www.w3.org/RDF/}}), $\top_D$
denotes the union of all domain values, $\INV{P}$ denotes the inverse
of $P$, $\SOMET{Q}$ denotes the objects related to by the role $Q$,
$\NOT$ denotes negation, $\DOMAIN{U}$ denotes the \emph{domain} of
$U$, i.e., the set of objects that $U$ relates to values, and
$\RANGE{U}$ denotes the \emph{range} of $U$, i.e., the set of values
related to objects by $U$.

A $\dlliteaid$ TBox $\T$ contains intensional assertions of three
types, namely inclusion assertions, functionality assertions, and
identification assertions \cite{CDLLR08b} (IDs). More precisely,
$\dlliteaid$ assertions are of the form:
\[
  \begin{array}{l@{\qquad}l}
    B \ISA C    & \mbox{\emph{concept inclusion assertion}}\\
    E \ISA T    & \mbox{\emph{value-domain inclusion
    assertion}}\\
    Q \ISA R    & \mbox{\emph{role inclusion assertion}}\\
    \funct{U}       & \mbox{\emph{attribute functionality assertion}}\\
    (id \; B\; \pi_1,...,\pi_n) & \mbox{\emph{identification
        assertions}}
  \end{array}
\]
In the identification assertions, $\pi$ denotes a \emph{path}, which
is an expression built according to the following syntax rule: $$\pi
\lora S ~\mid~ B? ~\mid~ \pi_1\circ\pi_2$$ 
where $S$ denotes an atomic role, the inverse of an atomic role, or an
atomic attribute, $\pi_1\circ\pi_2$ denotes the composition of the
paths $\pi_1$ and $\pi_2$, and $B?$, called \emph{test relation},
represents the identity relation on instances of the concept $B$. In
our logic, identification assertions are \emph{local}, i.e., at least
one $\pi_i \in \{\pi_1,...,\pi_n\}$ has length 1, i.e., it is an
atomic role, the inverse of an atomic role, or an atomic attribute. In
what follows, we only refer to IDs which are local.

A concept inclusion assertion expresses that a (basic) concept $B$ is
subsumed by a (general) concept $C$. Analogously for the other types
of inclusion assertions.  Inclusion assertions that do not contain
(resp. contain) the symbols '$\neg$' in the right-hand side are called
\emph{positive inclusions} (resp. \emph{negative
  inclusions}). Attribute functionality assertions are used to impose
that attributes are actually functions from objects to domain
values. Finally, an ID $ (id \; B\; \pi_1,...,\pi_n) $ asserts that
for any two different instances $a$,$b$ of $B$, there is at least on
$\pi_i$ such that $a$ and $b$ differ in the set of their
$\pi_i$-fillers. Note that IDs can be used to assert functionality of
roles. Specifically, the assertion $(id\;\SOMET{\INV{Q}}\;\INV{Q})$
imposes that $Q$ is functional.

The set of positive (resp., negative) inclusions in
$\T$ will be denoted by $\T^+$ (resp., $\T^-$), whereas the set of
identification assertions in $\T$ will be denoted by $\T_{id}$.


A $\dlliteaid$ ABox $\A$ is a finite set of assertions of the
form $A(a)$, $P(a,b)$, and $U(a,v)$, where $A$, $P$, and $U$
are as above,  $a$ and $b$ are object constants in $\S$, and
$v$ is a value constant in $\S$.
%
\begin{example}\label{es:kb}
  We consider a portion of the Formula One domain. We know that
  official drivers ($OD$) and test drivers ($TD$) are both team
  members ($TM$), and official drivers are not test drivers. Every
  team member is a member of ($mf$) a exactly one team
  ($\FOT$), and every team has at most one official driver. Finally,
  no race director ($RD$) is a member of a team. We also know that $s$
  is the official driver of team $t_1$, that $b$ is a test driver,
  and that $p$ is a team member. The corresponding $\dlliteaid$-KB
  $\K$ is:
  \begin{itemize}
  \item[$\T$:]  $OD \ISA TM$  $TD \ISA TM$  $OD \ISA \neg TD$  $RD \ISA \neg
  TM$ $TM \ISA \exists mf$ \\
  $TM \ISA \neg \FOT$  $\exists mf\ISA TM$  $\exists mf^-\ISA \FOT$
   $(id\;OD\;mf)$  $(id \; \FOT\; mf^-)$
\item[$\A$:]    $OD(s)$ \; $mf(s,t_1)$ \; $TD(b)$ \; $TM(p)$
\qed
  \end{itemize}

\end{example}

We conclude this section with a brief discussione on the
complexity of reasoning about a $\dlliteaid$-KB $\langle \T,
\A \rangle$. Satisfiability can be checked in polynomial time
with respect to $|\T \setminus \T_{id}|$ and $|\A|$, and in NP
with respect to $|\T_{id}|$. Moreover, if $\langle \T, \A
\rangle$ is satisfiable, then answering a query $q$ posed to
$\langle \T, \A \rangle$ can be done in polynomial time with
respect to $|\T|$ and $|\A|$, and in NP with respect to $|q|$.
Finally, $\clos{\T}{\A}$ can be computed in quadratic time
with respect to $|\T|$ and $|\A|$.

\section{WIDTIO approach to KB evolution in DLs}

In this section we first present our semantics for the evolution of DL
knowledge bases at the instance level, and then we provide a
comparison between our operator and other work in the literature.

\noindent \textbf{Semantics.~}
In what follows, $\L$ is a DL, and $\K = \langle \T,\A \rangle$ is a
satisfiable $\L$-KB. In other words, we do not consider the evolution
of unsatisfiable KBs. In addition, $F$ is a finite set of atomic ABox
assertions in $\L$.

The following definition specifies when a set of ABox assertions
``realizes'' the insertion or deletion of a set of ABox assertions
with respect to $\K = \langle \T,\A \rangle$.

\begin{definition}\label{def:accomplishing}
Let $\A'$ be a finite set of ABox assertions in $\L$.  Then,
we say that $\langle \T,\A' \rangle$ accomplishes the
insertion of $F$ into $\langle \T,\A \rangle$ if $\langle
\T,\A' \rangle$ is satisfiable, and $\langle \T,\A' \rangle
\models F$ (i.e., $F \subseteq \clos{\T}{\A'}$). Similarly,
$\langle \T,\A' \rangle$ accomplishes the deletion of $F$ from
$\langle \T,\A \rangle$ if $\langle \T,\A' \rangle$ is
satisfiable, and $\langle \T,\A' \rangle \not\models F$ (i.e.,
$F \not\subseteq \clos{\T}{\A'}$).
\end{definition}


Obviously, we are interested in KBs which accomplish the evolution of
a KB with a \emph{minimal change}. In order to formalize the
notion of \emph{minimal change}, we first need to provide some
definitions.

Let $\A_1$ and $\A_2$ be two finite sets of ABox assertions in
$\L$. Then, we say that $\langle \T,\A_1 \rangle$ has fewer
insertions than $\langle \T, \A_2 \rangle$ with respect to
$\langle \T,\A \rangle$ if $\clos{\T}{\A_1} \setminus
\clos{\T}{\A} \subset \clos{\T}{\A_2} \setminus
\clos{\T}{\A}$; and $\langle \T,\A_1 \rangle$ has fewer
deletions than $\langle \T,\A_2 \rangle$ with respect to
$\langle \T,\A \rangle$ if $\clos{\T}{\A} \setminus
\clos{\T}{\A_1} \subset \clos{\T}{\A} \setminus
\clos{\T}{\A_2}$. Also, we say that $\langle \T, \A_1 \rangle$
and $\langle \T, \A_2 \rangle$ have the same deletions with
respect to $\langle \T, \A \rangle$ if $\clos{\T}{\A}
\setminus \clos{\T}{\A_1} = \clos{\T}{\A} \setminus
\clos{\T}{\A_2}$.

\begin{definition}
  Let $\A_1$ and $\A_2$ be two finite sets of ABox assertions in
  $\L$. Then, $\langle \T,\A_1 \rangle$ has \emph{fewer changes} than
  $\langle \T, \A_2 \rangle$ with respect to $\langle \T,\A \rangle$
  if $\langle \T,\A_1 \rangle$ has fewer deletions than $\langle \T,
  \A_2 \rangle$ with respect to $\langle \T, \A \rangle$, or $\langle
  \T, \A_1 \rangle$ and $\langle \T, \A_2 \rangle$ have the same
  deletions with respect to $\langle \T, \A \rangle$, and $\langle
  \T,\A_1 \rangle$ has fewer insertions than $\langle \T, \A_2
  \rangle$ with respect to $\langle \T,\A \rangle$.
\end{definition}

Now that we have defined the relation of \emph{fewer changes} between
two KBs w.r.t. another one, we can define the notion of a KB which
accomplishes the insertion (resp. deletion) of a set of facts into
(resp. from) another KB minimally.

\begin{definition}\label{def:acco-min}
  The $\L$-KB $\langle \T,\A' \rangle$ accomplishes the insertion
  (deletion) of $F$ into (from) $\langle \T,\A \rangle$ minimally if
  $\langle \T,\A' \rangle$ accomplishes the insertion (deletion) of
  $F$ into (from) $\langle \T,\A \rangle$, and there is no
  $\L$-KB $\langle \T,\A'' \rangle$ that accomplishes the
  insertion (deletion) of $F$ into (from) $\langle \T,\A \rangle$, and
  has fewer changes than $\langle \T,\A' \rangle$ with respect to
  $\langle \T,\A \rangle$.
\end{definition}

With these notions in place, we can now define our evolution operator.

\begin{definition}\label{def:update-operator}
  Let $\U = \{ \langle \T,\A_1 \rangle,\ldots,\langle \T,\A_n \rangle
  \}$ be the set of all $\L$-KBs accomplishing the insertion
  (deletion) of $F$ into (from) $\langle \T,\A \rangle$ minimally, and
  let $\langle \T,\A' \rangle$ be an $\L$-KB. Then, $\langle \T,\A'
  \rangle$ is the result of changing $\langle \T,\A \rangle$ with the
  insertion (deletion) of $F$ if ~$(1)$~ $\U$ is empty, and $\langle
  \T,\clos{\T}{\A'} \rangle = \langle \T,\clos{\T}{\A} \rangle$, or
  ~$(2)$~ $\U$ is nonempty, and $\langle \T,\clos{\T}{\A'} \rangle =
  \langle \T,\bigcap_{1 \leq i \leq n }{\clos{\T}{\A_i}} \rangle$.
\end{definition}

It is immediate to verify that, up to logical equivalence, the result
of changing $\langle \T,\A \rangle$ with the insertion or the deletion
of $F$ is unique. In the rest of this paper, the result of changing
$\K = \langle \T,\A \rangle$ with the insertion (resp. deletion) of
$F$ according to our semantics will be denoted by $\K \oplus^\T_\cap
F$ (resp. $\K \ominus^\T_\cap F$). Notice that, by definition of our
operator, in the case where $F$ is inconsistent with $\T$, the result
of changing $\langle \T,\A \rangle$ with both the insertion and the
deletion of $F$ is logically equivalent to $\langle \T,\A \rangle$
itself.

\begin{example}\label{es:ins-del}
  Consider the $\dlliteaid$ KB $\K$ of the Example \ref{es:kb}, and
  suppose that $p$ becomes now a race director, and $b$ becomes the
  new official driver ofq team $t_1$. To reflect this new information,
  we change $\K$ with the insertion of $F_1=\{
  RD(p),OD(b),mf(b,t_1)\}$. Since the TBox implies that a race
  director cannot be a team member, $RD(p)$ contradicts $TM(p)$. Also,
  since every team has at most one official driver, $OD(b)$ and
  $mf(b,t_1)$ contradict $mf(s,t)$. According to Definition
  \ref{def:acco-min}, the KBs accomplishing the insertion of $F_1$
  into $\K$ minimally are:
\begin{itemize}
\item[]$\K_1$ = $\langle \T,$ $\{RD(p)$,$OD(b)$,$mf(b,t_1)$,$TM(s)$,$mf(s,t_1)\}\rangle$
\item[]$\K_2$ = $\langle \T,$ $\{RD(p)$,$OD(b)$,$mf(b,t_1)$,$TM(s)$,$OD(s)\}\rangle$
\end{itemize}
\noindent Thus, $\K \oplus^\T_\cap F_1$ is:
\begin{itemize}
\item[]$\K_3$ = $\langle \T,$ $\{RD(p)$,$OD(b)$,$mf(b,t_1)$,$TM(s)\}\rangle$.
\end{itemize}
\noindent Now, suppose that we do not know anymore whether $b$ is a
member of $t_1$, and, even more, whether $b$ is a team member at
all. Then, we change $\K_3$ with the deletion of
$F_2=\{TM(b),mf(b,t_1)\}$, thus obtaining
\begin{itemize}
\item[]$\K_3 \oplus^\T_\cap F_2$ = $\langle \T,$
  $\{RD(p)$,$TM(s)$,$OD(b)\}\rangle$. \qed
\end{itemize}
\end{example}

\noindent \textbf{Comparison with related work.} \;
We mentioned in the introduction several model-based approaches to DL
KB evolution, and noticed that they all suffer from the expressibility
problem.  This problem is also shared by \cite{WangWT10}, that uses
\emph{features} instead of models, and proposes the notion of
approximation to cope with the expressibility problem, similarly to
\cite{DLPR09}. 

Related to our proposal are several formula-based approaches
proposed in the literature. We already pointed out that our
proposal is inspired by \cite{FaUV83}, although the problem
studied in \cite{FaUV83} is evolution in propositional logic,
whereas the context dealt with in our work is instance-level
evolution in DLs. Perhaps, the closest approach to the one
proposed in this paper is that reported in \cite{CKNZ10b},
where formula-based evolution (actually, insertion) of
$\dllite$ KBs is studied. The main difference with our work is
that we base our semantics on the WIDTIO principles, and
therefore we compute the intersection of all KBs accomplishing
the change minimally. Conversely, in the \emph{bold} semantics
discussed in \cite{CKNZ10b}, the result of the change is
chosen non-deterministically among the KBs accomplishing the
change minimally. Another difference is that while
\cite{CKNZ10b} addresses the issue of evolution of both the
TBox and the ABox, we only deal with the case of fixed TBox
(in the terminology of \cite{CKNZ10b}, this corresponds to
keep the TBox \emph{protected}). It is interesting to observe
that the specific DL considered in \cite{CKNZ10b} is
$\dllitefr$, and for this logic, exactly one KB accomplishes
the insertion of a set of ABox assertions minimally. It
follows that for instance-level insertion, their bold
semantics coincides with ours. On the other hand, the presence
of identification assertions in $\dlliteaid$ changes the
picture considerably, since with such assertions in the TBox,
many KBs may exist accomplishing the insertion minimally. In
this case, the two approaches are indeed different. Finally,
\cite{CKNZ10b} proposes a variant of the bold semantics,
called \emph{careful semantics}, for instance-level insertion
in $\dllitefr$. Intuitively, such a semantics aims at
disregarding knowledge that is entailed neither by the
original KB, nor by the set of newly asserted facts.  Although
such principle is interesting, we believe that the careful
semantics is too drastic, as it tends to eliminate too many
information from the original KB as shown in the following
example.
\begin{example}
  Consider the KB $\K$ of the Example \ref{es:kb}, and suppose that we
  $c$ is now a member of a formula one team, which means changing $\K$
  with the insertion of $TM(c)$. Notice that such a new fact does not
  contradict any information in $\K$. Therefore, in our approach, the
  result of the insertion is $\langle \T,$ $\{OD(s)$, $mf(s,t_1)$,
  $TD(b)$, $TM(p)$, $TM(c)$ $\}\rangle$. Conversely, one can verify
  that the result under the careful semantics is $\langle \T,$
  $\{OD(s)$, $mf(s,t_1)$, $TM(c)$ $\}\rangle$, thus loosing both the
  information that $b$ is a test driver, and the information that $p$
  is a team member. \qed
\end{example}

Finally, we point out that, to our knowledge, the evolution operator
presented in this work is the first tractable evolution operator based
on the WIDTIO principle.

\section{Deletion in $\dlliteaid$}

We study deletion under the assumption that the DL language $\L$ is
$\dlliteaid$. Thus, we refer to a $\dlliteaid$-KB $\K = \langle \T,\A
\rangle$, and we address the problem of changing $\K$ with the
deletion of a finite set $F$ of ABox assertions. We assume that both
$\langle \T,\A \rangle$ and $\langle \T,F \rangle$ are satisfiable.

The following theorem specifies when a $\dlliteaid$-KB accomplishes
the deletion of $F$ from $\langle \T,\A \rangle$ minimally.

\begin{theorem}\label{thm-5-1}
  $\langle \T,\A' \rangle$ accomplishes the deletion of $F$ from
  $\langle \T,\A \rangle$ minimally if and only if $\clos{\T}{\A'}$ is
  a maximal $\T$-closed subset of $\clos{\T}{\A}$ such that $F \not
  \subseteq \clos{\T}{\A'}$.
\end{theorem}

We now consider the case where the set $F$ is constituted by just one
assertion $f$. By exploiting Theorem~\ref{thm-5-1}, it is easy to
conclude that there is exactly one KB accomplishing the deletion of
$\{ f \}$ from a given KB.

\begin{theorem}\label{thm-5-2}
  Let $f$ be an ABox assertion. Up to logical equivalence, there is
  exactly one KB of the form $\langle \T, \A' \rangle$ that
  accomplishes the deletion of $\{ f \}$ from $\langle \T, \A \rangle$
  minimally, and such KB can be computed in polynomial time with
  respect to $|\T|$ and $|\A|$.
\end{theorem}

\begin{proofsk}
  The proof is based on the fact that $\langle \T, \A \setminus
  \textsf{Subsumee}_\K(f) \rangle$ is the unique maximal $\T$-closed
  subset $\A'$ of $\clos{\T}{\A}$ such that $\{ f \} \not \subseteq
  \clos{\T}{\A'}$.  
\end{proofsk}

Let us now consider the case of arbitrary $F = \{
f_1,\ldots,f_m\}$. Suppose that, for every $1 \leq i \leq m$, $\langle
\T,\A_i \rangle$ accomplishes the deletion of $\{f_i\}$ from $\langle
\T,\A \rangle$ minimally. One might wonder whether the set $\Gamma_1$
of all KBs accomplishing the deletion of $F$ from $\langle \T,\A
\rangle$ minimally coincides (modulo logical equivalence) with
$\Gamma_2= \{ \langle \T,\A_1 \rangle, \ldots \langle \T,\A_m
\rangle\}$. The next theorem tells us that one direction is indeed
valid: for each KB $\K_1 \in \Gamma_1$ there exists a KB $\K_2 \in
\Gamma_2$ such that $\Mod{\K_1}=\Mod{\K_2}$.

\begin{theorem}\label{thm-5-3}
  If $\langle \T,\A' \rangle$ accomplishes the deletion of $\{
  f_1,\ldots,f_m \}$ from $\langle \T,\A \rangle$ minimally, then
  there exists $i \in \{1..m \}$ such that $\langle \T,\A' \rangle$
  accomplishes the deletion of $f_i$ from $\langle \T,\A \rangle$
  minimally.
\end{theorem}
However, the following example shows that the other direction does not
hold: there may exist a $\K_2 \in \Gamma_2$ that is not logically
equivalent to any $\K_1 \in \Gamma_1$.

\begin{example}
  Let $\T$ be $\{ B \ISA C, C \ISA D, E \ISA D \}$, let $\A$ be $\{
  B(a), E(a)\}$, and let $F$ be $\{ C(a), D(a)\}$. It is easy to see
  that the deletion of $D(a)$ from $\langle \T,\A \rangle$ is
  accomplished minimally by $\langle \T,\emptyset \rangle$, while the
  deletion of $C(a)$ from $\langle \T,\A \rangle$ is accomplished
  minimally by $\langle \T,\{ E(a) \} \rangle$. Therefore, in this
  case $\Gamma_2 = \{ \langle \T,\emptyset \rangle, \langle \T,\{ E(a)
  \} \rangle \}$. Also, one can verify that $\langle \T,\{ E(a) \}
  \rangle$ is the only (up to logical equivalence) KB accomplishing
  the deletion of $F$ minimally, i.e., $\Gamma_1 = \{\langle \T,\{
  E(a) \} \rangle \}$. Thus, there is a KB in $\Gamma_2$, namely
  $\langle \T,\emptyset \rangle$, that is not logically equivalent to
  any KB in $\Gamma_1$. \qed
\end{example}

Note that the above example also shows that deleting $F$ is not
equivalent to iteratively deleting all atoms in $F$.

The next theorem characterizes when a given $\langle \T,\A_i \rangle
\in \Gamma_2$ accomplishes the deletion of $F$ minimally.

\begin{theorem}\label{thm-5-4}
  Let $F = \{ f_1,\ldots,f_m\}$, and, for every $1 \leq i \leq m$, let
  $\langle \T,\A_i \rangle$ accomplish the deletion of $\{f_i\}$ from
  $\langle \T,\A \rangle$ minimally. Then, $\langle \T,\A_j \rangle$,
  where $j \in \{1..m\}$, accomplishes the deletion of $F$ from
  $\langle \T,\A \rangle$ minimally if and only if there is no $h \in
  \{1..m\}$ such that $h \neq j$, and $\langle \T, \{f_h\} \rangle
  \models f_j$.
\end{theorem}

\begin{proofsk} We first show that $(\alpha)$ $\langle
  \T,\A_j \rangle$, where $j \in \{1..m\}$, accomplishes the deletion
  of $F$ from $\langle \T,\A \rangle$ minimally if and only if there
  is no $h \in \{1..m\}$ such that $\A_j \subset \A_h$, and then show
  that $(\beta)$ $\A_j \subset \A_h$ if and only if $h \neq j$, and
  $\langle \T, \{f_h\} \rangle \models f_j$. 
\end{proofsk}

By exploiting Theorems~\ref{thm-5-2}, \ref{thm-5-3}, and
\ref{thm-5-4}, we can directly prove that $\K \ominus^\T_\cap F$ can
be computed by the algorithm \textit{$ComputeDeletion$} below. It is
easy to see that the time complexity of the algorithm is $O( |\T|^2
\times |F|^2 + |\A|^2)$.

%
\SetAlFnt{\small}
\begin{algorithm}[htbp]
  \KwIn{a satisfiable $\dlliteaid$ KB $\K = \tup{\T,\A}$, a finite set
    of ABox assertions $F$ such that $\langle \T, F
    \rangle$ is satisfiable}
 \KwOut{a $\dlliteaid$ KB.}
\Begin{
$F'\leftarrow F$;\\
\ForEach{$f_i\in F'$ and $f_j \in F$ such that $i \neq j$}
{\lIf{$\tup{\T,\{f_j\}} \models f_i$} {$F' \leftarrow F'
\setminus \{f_i\}$\\}}
\Return{$\tup{\T,\clos{\T}{\A}\setminus \{\alpha \in \textsf{Subsumee}_\K(f) ~|~ f \in F' \}}$};\\
} \caption{Algorithm
\textit{$ComputeDeletion(\tup{\T,\A},F)$}}
\label{alg:computeDEL}
\end{algorithm}
\SetAlFnt{\normalsize}
%


\begin{theorem}
  $ComputeDeletion(\tup{\T,\A},F)$ terminates, and computes $\tup{\T,\A}
  \ominus^\T_\cap F$ in polynomial time with respect to $|\T|$, $|\A|$
  and $|F|$.
\end{theorem}

\section{Insertion in $\dlliteaid$}

We refer to a $\dlliteaid$-KB $\K = \langle \T,\A \rangle$, and we
address the problem of changing $\K$ with the insertion of a finite
set $F$ of ABox assertions. As in the previous section, we assume that
both $\langle \T,\A \rangle$ and $\langle \T,F \rangle$ are
satisfiable. The main problem to be faced with insertion is described
by the following observation. 

Suppose that $\T$ contains $n$ identification assertions with at least
two atoms that become simoultaneously violated with the insertion of a
single ABox assertion $f$ into $\langle \T,\A \rangle$, and such that
every choice of retracting one of such atoms yields a maximal subset
of $\clos{\T}{\A}$ that is $\T$-consistent with $f$. Obviously, there
are at least $2^n$ such maximal subsets. What the above example shows
is that, given $f$, there can be an exponential number of maximal
subsets $\A'$ of $\clos{\T}{\A}$ such that $\langle \T,\A' \cup \{f\}
\rangle$ is satisfiable.  Note that this cannot happen in those DLs of
the $\dllite$ family which do not admit the use of identification
assertions (such as the DL studied in \cite{CKNZ10b}). Indeed, in such
logic, there is always one maximal subset of $\clos{\T}{\A}$ that is
consistent with a set $F$ of ABox assertions.

It follows from the above observation that building all maximal
subsets of $\langle \T,\A \rangle$ which are $\T$-consistent with $F$,
and then computing their intersection is computationally
costly. Fortunately, we show in the following that we can compute $\K
\oplus^\T_\cap F$ without computing all maximal consistent subsets
of $\langle \T,\A \rangle$ with $F$.

To describe our method, we need some preliminary notions. A set $V$ of
facts is called a \emph{$\T$-violation set for $t \in \T \setminus
  \T^+$} if $\langle \T^+ \cup \{t\},V \rangle$ is unsatisfiable,
while for every proper subset $V'$ of $V$, $\langle \T^+ \cup \{t\},V'
\rangle$ is satisfiable. Any set $V$ of facts that is a $\T$-violation
set for a $t \in \T \setminus \T^+$ is simply called a
\emph{$\T$-violation set}.

\begin{theorem}\label{thm-6-1}
  Let $\langle \T, \A \rangle$ be a satisfiable $\dlliteaid$-KB, and
  let $\alpha$ be an ABox assertion such that $\langle \T, \{\alpha\}
  \rangle$ is satisfiable. If $\langle \T, \A \cup \{ \alpha \}
  \rangle$ is unsatisfiable, then there is a $\T$-violation set $V$ in
  $\clos{\T}{\A \cup \{\alpha\}}$ such that $(i)$ $V$ contains
  $\alpha$, and $(ii)$ $(V \setminus \{ \alpha \}) \subseteq
  \clos{\T}{\A}$.
\end{theorem}
\begin{proofsk}
  We first show that, if $\langle \T, \A \cup \{ \alpha \} \rangle$ is
  unsatisfiable, then there is a TBox assertion $t$ in $\T \setminus
  \T^+$ such that $\langle \T^+, \A \cup \{ \alpha \} \rangle \models
  q^t$, where $q^t$ is the boolean query corresponding to the negation
  of $t$. This implies that there is a query $q'$ in the
  $\T$-expansion of $q^t$ that evaluates true on $\clos{\T}{\A \cup
    \{\alpha\}}$, i.e., that forms a $\T$-violation set for $t$ in
  $\clos{\T}{\A \cup \{\alpha\}}$. Now suppose that, for every $t \in
  \T \setminus \T^+$, and for every $\T$-violation set $V$ in
  $\clos{\T}{\A \cup \{\alpha\}}$, $V$ does not contain $\alpha$. This
  means that either $(i)$ there is no $\T$-violation set in
  $\clos{\T}{\A \cup \{\alpha\}}$, or $(ii)$ all $\T$-violation sets
  in $\clos{\T}{\A \cup \{\alpha\}}$ do not contain $\alpha$. Btoh
  cases lead to a contradiction, and, therefore, we conclude that
  there is a $\T$-violation set $V$ in $\clos{\T}{\A \cup \{\alpha\}}$
  such that $V$ contains $\alpha$. Finally, since $\langle \T,
  \{\alpha\} \rangle$ is satisfiable, it is immediate to verify that
  $(V \setminus \{ \alpha \}) \subseteq \clos{\T}{\A}$.
\end{proofsk}

The next theorem is the key to our solution.

\begin{theorem}\label{thm-6-2}
  Let $\alpha$ be an atom such that $\alpha \in
  \clos{\T}{\A}\setminus\clos{\T}{F}$. There exists a maximal subset
  $\Sigma$ of $\clos{\T}{\A}$ such that $\langle \T,\Sigma \cup F
  \rangle$ is satisfiable and $\Sigma$ does not contain $\alpha$ if
  and only if there is a $\T$-violation set $V$ in $\clos{\T}{\A} \cup
  \clos{\T}{F}$ such that $\alpha \in V$, and $\langle \T,F \cup (V
  \setminus \{ \alpha \})\rangle$ is satisfiable.
\end{theorem}
\begin{proofsk} $(\Rightarrow)$ Suppose that there is a $\T$-violation
  set $V$ in $\clos{\T}{\A} \cup \clos{\T}{F}$ such that $\alpha \in
  V$ and $\langle \T,F \cup (V \setminus \{ \alpha \}) \rangle$ is
  satisfiable. Since $\langle \T,F \cup (V \setminus \{ \alpha
  \})\rangle$ is satisfiable, the set of maximal subsets $\Sigma$ of
  $\clos{\T}{\A}$ such that $\langle \T,\Sigma \cup F \cup (V
  \setminus \{ \alpha \})\rangle$ is satisfiable is
  non-empty. Consider any $\Sigma$ in such a set, i.e., assume that
  $\Sigma$ is a maximal subset of $\clos{\T}{\A}$ such that $\langle
  \T, \Sigma \cup F \cup (V \setminus \{ \alpha \})\rangle$ is
  satisfiable. It can be shown that (1) $\Sigma$ does not contain
  $\alpha$, and (2) $\Sigma$ is a maximal subset of $\clos{\T}{\A}$
  such that $\langle \T,\Sigma \cup F \rangle$ is satisfiable.

$(\Leftarrow)$ Suppose that there is no $\T$-violation set $V$ in
$\clos{\T}{\A} \cup \clos{\T}{F}$ such that $\alpha \in V$ and
$\langle \T,F \cup (V \setminus \{ \alpha \}) \rangle$ is
satisfiable. We show that every maximal subset $\Sigma'$ of
$\clos{\T}{\A}$ such that $\langle \T, \Sigma' \cup F \rangle$ is
satisfiable contains $\alpha$, by showing that, if $\Sigma$ is a
subset of $\clos{\T}{\A}$ such that $\langle \T, \Sigma \cup F
\rangle$ is satisfiable, then $\langle \T, \Sigma \cup F \cup
\{\alpha\} \rangle$ is also satisfiable. Indeed, assume by way of
contradiction that $\langle \T, \Sigma \cup F \cup \{\alpha\} \rangle$
is unsatisfiable.  Note that $\alpha \in \clos{\T}{\A}$, and, since
$\langle \T, \A \rangle$ is satisfiable, $\langle \T, \{ \alpha \}
\rangle$ is also satisfiable. We can therefore apply
theorem~\ref{thm-6-1}, and conclude that there is a $\T$-violation set
$V$ in $\clos{\T}{F \cup \Sigma \cup \{\alpha\}}$ such that (1) $V$
contains $\alpha$, (2) $(V \setminus \{ \alpha \}) \subseteq
\clos{\T}{F \cup\Sigma}$.  Now, since $(V \setminus \{ \alpha \})
\subseteq \clos{\T}{F \cup\Sigma}$, and $\langle \T, \Sigma \cup F
\rangle$ is satisfiable, it follows that $\langle \T, F \cup (V
\setminus \{ \alpha \}) \rangle$ is satisfiable.  This implies that
there is a $\T$-violation set $V$ in $\clos{\T}{F \cup \Sigma \cup
  \{\alpha\}} \subseteq \clos{\T}{\A} \cup \clos{\T}{F}$ such that
$\alpha \in V$ and $\langle \T, F \cup (V \setminus \{ \alpha \})
\rangle$ is satisfiable, which is a contradiction. 
\end{proofsk}

Theorems~\ref{thm-6-1} and \ref{thm-6-2} allow us to prove that $\K
\oplus^\T_\cap F$ can be computed by the algorithm
\textit{$ComputeInsertion$} below.
\SetAlFnt{\footnotesize}
\begin{algorithm}[htbp]
  \KwIn{a satisfiable $\dlliteaid$ KB $\K = \tup{\T,\A}$, a finite set
    of ABox assertions $F$ such that $\langle \T, F
    \rangle$ is satisfiable}
 \KwOut{a $\dlliteaid$ KB.}
\Begin{ $F' = \emptyset$;\\
\ForEach{$\alpha \in \clos{\T}{\A} \setminus \clos{\T}{F}$} {
    \uIf{$\exists$ a $\T$-violation set $V$ in $\clos{\T}{\A} \cup \clos{\T}{F}$ s.t. $\alpha \in V$ and
     $\langle \T,F \cup (V \setminus \{ \alpha \}) \rangle$ is
     satisfiable}
    {
     $F'\leftarrow F' \cup \{\alpha\}$\\
    }
 }
\Return{$\tup{\T,F \cup \clos{\T}{\A} \setminus F'}$};\\
} \caption{Algorithm
\textit{$ComputeInsertion(\tup{\T,\A},\F)$}}
\label{alg:computeINS}
\end{algorithm}

Algorithm $ComputeInsertion$ requires to compute all $\T$-violation
sets in $\clos{\T}{\A} \cup \clos{\T}{F}$. It can be shown that this
can be done by computing the results of suitable conjunctive queries
posed to $\clos{\T}{\A} \cup \clos{\T}{F}$. Such queries are built out
of the negative inclusion assertions and the identification assertions
$\T_{id}$ in $\T$, and essentially look for tuples that satisfy the
negation of such assertions. From this observation, one can derive the
following theorem.

\begin{theorem}
  $ComputeInsertion(\tup{\T,\A},\F)$ terminates, and computes
$\tup{\T,\A} \oplus^\T_\cap F$ in polynomial time with respect to $|\T
\setminus \T_{id}|$, $|\A|$, and $|F|$, and in NP with respect to
$|\T_{id}|$.
\end{theorem}

It can also be shown that the problem of checking for the existence of
$\T$-violation sets in a set of ABox assertions is NP-complete with
respect to $|\T_{id}|$.

\section{Conclusions}

We have illustrated a WIDTIO approach to instance-level evolution in
DL, and we have presented algorithms for the case of $\dlliteaid$. We
plan to continue our work along several directions. First, we will
extend the algorithms to the case where the KB contains denial
constraints, which are constraints that can be added to $\dlliteaid$
without changing the complexity of all reasoning tasks. The extension
is based on the fact that denial constraints behave similarly to
identification assertions with respect to KB evolution. Also, we aim
at extending our approach to the problem of evolution of the whole KB,
as opposed to the ABox only. Finally, we will add the notion of
protected part to our approach, to model situations where one wants to
prevent changes on specific parts of the KB when applying insertions
or deletions.

\end{document}